\documentclass{esannV2}
\usepackage{graphicx}
\usepackage[latin1]{inputenc}
\usepackage{amssymb,amsmath,array,amsthm,mathtools}
\usepackage{booktabs}
\usepackage{enumitem}
\usepackage{watermark}
\watermark{\thispageheading{\sffamily Proceedings of the 29th European Symposium on Artificial Neural Networks, Computational Intelligence and Machine Learning (ESANN 2021), pp. 99-104 doi:10.14428/esann/2021.ES2021-70}}
\theoremstyle{definition}
\newtheorem{theorem}{Theorem}

\newtheorem{proposition}[theorem]{Proposition}

\theoremstyle{remark}

\newcommand{\norm}[1]{\left\lVert #1 \right\rVert}
\newcommand{\vect}[1]{\mathbf{#1}}
\newcommand{\sequ}[1]{\boldsymbol{#1}}
\newcommand{\matx}[1]{\mathbf{#1}}
\newcommand{\eqdef}{\stackrel{\mathclap{\normalfont\mbox{def}}}{=}}

\newcommand{\gmean}[1]{\left\lceil #1 \right\rfloor}

\begin{document}
\title{Dynamic Graph Echo State Networks}

\author{Domenico Tortorella and Alessio Micheli
\vspace{.3cm}\\
University of Pisa - Department of Computer Science \\
Largo B. Pontecorvo 3, 56127 Pisa - Italy
}

\maketitle

\begin{abstract}
Dynamic temporal graphs represent evolving relations between entities, e.g. interactions between social network users or infection spreading.
We propose an extension of graph echo state networks for the efficient processing of dynamic temporal graphs, with a sufficient condition for their echo state property, and an experimental analysis of reservoir layout impact.
Compared to temporal graph kernels that need to hold the entire history of vertex interactions, our model provides a vector encoding for the dynamic graph that is updated at each time-step without requiring training.
Experiments show accuracy comparable to approximate temporal graph kernels on twelve dissemination process classification tasks.
\end{abstract}

\section{Introduction}
\begin{figure}[h!]
	\centering
	\includegraphics[scale=0.16]{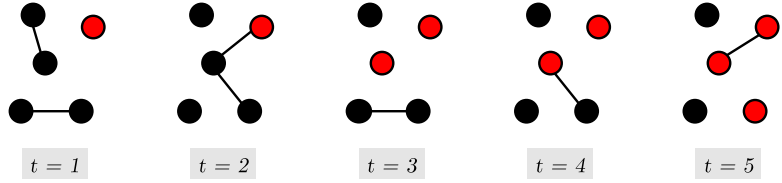}
	\caption{The spreading of an infection, an example of dissemination process. Infected vertices (in red, $\vect{u}_t(v) = 1$) contaminate susceptible vertices (in black, $\vect{u}_t(v) = 0$) by interacting with them at time-step $t$ (temporal edges).}
	\label{fig:dissemination-process}
\end{figure}

Graphs are relevant in modelling entities and relations between them, e.g. atoms and bonds in a molecule, or paper citation networks.
A plethora of machine learning models able to treat directly these data structures has been proposed and successfully employed for classification and regression tasks, both on graphs and on vertices \cite{Bacciu2020}.
However, some relations evolve through time. People spreading a disease by interacting at certain instants, or users sharing posts on a social network, are just two examples of \emph{dissemination processes} (Fig. \ref{fig:dissemination-process}).
Many approaches have been recently proposed for learning representation of dynamic graphs \cite{Kazemi2020}.
In particular, temporal graph kernels \cite{Oettershagen2020} extend classic kernels employed for static graphs by transforming dynamic graphs into equivalent static ones; spatio-temporal graph convolutional networks \cite{Wu2021} deal directly with dynamic graphs, but require end-to-end training.
On the other hand, efficient reservoir approaches such as echo state networks (ESN), which embed data structures without requiring recurrent weights training, have been successfully applied to classification and regression tasks on static graphs \cite{Gallicchio2010,Gallicchio2020}.

In section \ref{sec:model} we present DynGESN\footnote{Code available at \texttt{github.com/dtortorella/dyngraphesn}, based on the original GraphESN implementation by C. Gallicchio.}, an extension of GraphESN for dynamic temporal graphs, along with a sufficient condition for the echo state property.
In section \ref{sec:experiments} we evaluate our model on twelve classification tasks, analysing the impact of reservoir layout on accuracy and comparing performances with temporal graph kernels.

\section{Model}\label{sec:model}
We define a \emph{dynamic} graph $\mathcal{G}$ as a pair $(\mathcal{V}, \mathcal{E})$, where $\mathcal{V}$ is the set of vertices, and $\mathcal{E} = \{(u, v, t) \,|\, u, v \in \mathcal{V}, t \in 1..T\}$ is the set of edges $\textcircled{\textit{u}} \mkern-6mu \to \mkern-6mu \textcircled{\textit{v}}$ between a pair of vertices at a time-step $t$.
The graph $\mathcal{G}$ can be characterized as \emph{undirected} if $(u, v, t) \in \mathcal{E} \Rightarrow (v, u, t) \in \mathcal{E}$, and as \emph{static} if $(u, v, t) \in \mathcal{E} \Rightarrow (u, v, t') \in \mathcal{E} \,\forall t, t'$ (i.e., constant edge set).
We also define $\mathcal{E}_t = \{(u, v) \,|\, (u, v, t) \in \mathcal{E}\}$, the neighbourhood $\mathcal{N}_t(v) = \{u \in \mathcal{V} \,|\, (u, v) \in \mathcal{E}_t\}$, and $\matx{A}_t$ as the adjacency matrix at time-step $t$.
Finally, we associate at each vertex $v$ a label sequence $\vect{u}_t(v) \in \mathbb{R}^U$ (e.g. $\vect{u}_t(v) \in \{0, 1\}$ in Fig. \ref{fig:dissemination-process}).

The ESNs are a particular class of recurrent neural networks (RNN) in which the recurrent weights are randomly initialized under certain conditions (see later) and kept fixed, while only a memoryless readout layer is trained.
Formally, an ESN is characterized as an input-driven dynamical system governed by a transition function $\vect{x}_t = F(\vect{u}_t, \vect{x}_{t-1})$ with states $\vect{x}_t$ belonging to a compact subset $\mathcal{X}$ of $\mathbb{R}^H$, and $\vect{u}_t \in \mathbb{R}^U$ being the input at time $t$.
An input sequence $\sequ{u} = [\vect{u}_1 .. \vect{u}_T]$ is encoded into the embedding space $\mathcal{X}$ by applying iteratively $F$ from $t = 1$ to the end,
with $\vect{x}_0 \in \mathcal{X}$ as initial state for $t = 0$.

We extend the deep ESN reservoir model for vector sequences \cite{Gallicchio2017} to dynamic temporal graphs by having each recurrent layer $\ell \in 1..L$ compute new vertex features $\vect{x}^{(\ell)}_t(v) \in \mathcal{X} = [-1,+1]^H$ by the transition function $F^{(\ell)}(\vect{u}_t, \matx{A}_t, \vect{x}_{t-1})$ defined vertex-wise as
\begin{equation}\label{eq:dyngrahpesn}
	\vect{x}^{(\ell)}_t(v) = \gamma_{\ell} \, \tanh \left(\matx{W}_{\mathrm{in}}^{(\ell)} \, \vect{x}^{(\ell-1)}_t(v) + \displaystyle\sum_{v' \in \mathcal{N}_t(v)} \matx{\hat{W}}^{(\ell)} \, \vect{x}_{t-1}^{(\ell)}(v') \right) + (1 - \gamma_{\ell}) \, \vect{x}_{t-1}^{(\ell)}(v)
\end{equation}
with $0 < \gamma_{\ell} \leq 1$ as leakage constant, $\vect{x}_t^{(0)}(v) = \vect{u}_t(v)$ for layer $\ell = 1$, and initial state $\vect{x}^{(\ell)}_0(v) = \vect{0}$.
Both input weights $\matx{W}_{\mathrm{in}}^{(\ell)}$ and recurrent weights $\matx{\hat{W}}^{(\ell)}$ are randomly initialized.
The embedding of a graph $\mathcal{G}$ is then given by the pooled vertex features of the final state for each layer
\begin{equation}\label{eq:pooling}
	\vect{X}_{\mathcal{G}} = \begin{bmatrix}
		\sum_{v \in \mathcal{V}} \vect{x}_T^{(1)}(v) & 
		\dots &
		\sum_{v \in \mathcal{V}} \vect{x}_T^{(L)}(v)
	\end{bmatrix} \in \mathbb{R}^{H L}
\end{equation}
and can be used to perform regression or classification, e.g. by training a linear readout
\begin{equation}\label{eq:readout}
	\vect{y} = \matx{W}_\mathrm{out} \, \vect{X}_{\mathcal{G}} + \vect{b}
\end{equation}
by ridge regression, or by training a support vector machine (SVM).

Equation \eqref{eq:dyngrahpesn} resembles closely the transition function of GraphESN \cite{Gallicchio2010}, except for the use of temporal neighbourhood $\mathcal{N}_t$ due to the change in graph connectivity between time-steps.
However, they operate rather differently: GraphESN processes static graphs with constant vertex labels by iterating its transition function until the global state converges to a fixed point ($t$ tracking iterations in this case), while ours acts analogously to ESN for finite sequences (in our case a sequence of $T$ graphs over time $t$).
Furthermore, the deep GraphESN \cite{Gallicchio2020} waits for convergence in the previous layer to pass the final state as input to the subsequent layer, while our model works more similarly to deep ESN for sequences, i.e. passing a state to the subsequent layer at each time-step \cite{Gallicchio2017}.

ESN are able to provide meaningful encoding of sequences thanks to the echo state property (ESP) \cite{Gallicchio2017}, which ensures that perturbations in the initial state are `washed out' in the long term.
Furthermore, the contractivity of $F$, i.e. $\norm{F(\vect{u}, \vect{x}) - F(\vect{u}, \vect{x}')} \leq C \norm{\vect{x} - \vect{x}'},\, C < 1$ (reducing embedding state distance under same input), ensures that the embedding space of sequences has a suffix-based organization \cite{Gallicchio2011}.

\begin{proposition}[DynGESN ESP]\label{pr:dyngesn-esp}
The transition function defined in equation \eqref{eq:dyngrahpesn} is contractive with constant $C_t = (1 - \gamma^{(\ell)}) + \gamma^{(\ell)} \, \lVert\matx{\hat{W}^{(\ell)}}\rVert\, \norm{\matx{A}_t}$ for each layer $\ell$.
Therefore, a sufficient condition for the ESP to hold is
\begin{equation}\label{eq:dyngraphesn-esp}
	\norm{\matx{\hat{W}^{(\ell)}}} < \frac{1}{\gmean{\alpha^{*}_t}} \quad \text{for all layers} \; \ell \in 1..L,
\end{equation}
where $\alpha^{*}_t = \norm{\matx{A}_t}$ is the maximum eigenvalue/singular value of the adjacency matrix $\matx{A}_t$, and $\gmean{\alpha^{*}_t}$ is the geometric mean $\gmean{\alpha^{*}_t}\; \eqdef \; \limsup_{T \to \infty} \sqrt[T]{\prod_{1 \leq t \leq T} \alpha^{*}_t}$.
\end{proposition}
The proof works out analogously to the sufficient condition for ESN in \cite{Gallicchio2017}, and is omitted for brevity.
Interestingly, Proposition \ref{pr:dyngesn-esp} provides a weaker ESP sufficient condition for GraphESN \cite{Gallicchio2020}: in case of static graphs $\gmean{\alpha^{*}_t} = \alpha^{*} \leq k^{*}$, where $k^{*}$ is the maximum vertex degree, and $\alpha^{*} = k^{*}$ for complete graphs.

\section{Experiments and discussion}\label{sec:experiments}
We evaluate our model for dynamic graph binary classification on twelve datasets developed by Oettershagen et al. \cite{Oettershagen2020}, where dissemination processes based on the \emph{susceptible-infected} (SI) epidemic model have been simulated on six different real-world social interaction datasets.
In a SI model vertices are labelled either \emph{susceptible} or \emph{infected}, switching label from former to latter with fixed probability $p$ at each time-step when they are directly linked to an infected vertex.

The two classes of the first six tasks (\textsf{ct1}) are dynamic graphs that follow a SI process with probability $p = 0.5$, and dynamic graphs whose vertices randomly switch from susceptible to infected.
In the second six tasks (\textsf{ct2}) classifiers have to discriminate between two SI dissemination processes with contagion probability $p = 0.2$ and $p = 0.8$.
We refer to \cite{Oettershagen2020} for further details.

Since our model depends on random weights initialization, we evaluated classification accuracies by averaging on $200$ bootstraps with $90\%$-$10\%$ training/test splits instead of a $10$-fold cross-validation.
Reservoir weights are randomly generated following an uniform distribution, with recurrent weights rescaled to $\lVert\matx{\hat{W}^{(\ell)}}\rVert\ = 0.9 \, \alpha_{\mathrm{mean}}$ in order to satisfy Proposition \ref{pr:dyngesn-esp}, $\alpha_{\mathrm{mean}}$ being the average $\gmean{\alpha^{*}_t}$ on each dataset; we fixed $\gamma_{\ell} = 0.1$.
A linear readout \eqref{eq:readout} is trained by ridge regression with regularization $\lambda = 10^{-3}$.

\begin{figure}[h!]
	\centering
	\includegraphics[scale=0.65, trim=4cm 0.4cm 4cm 0.2cm]{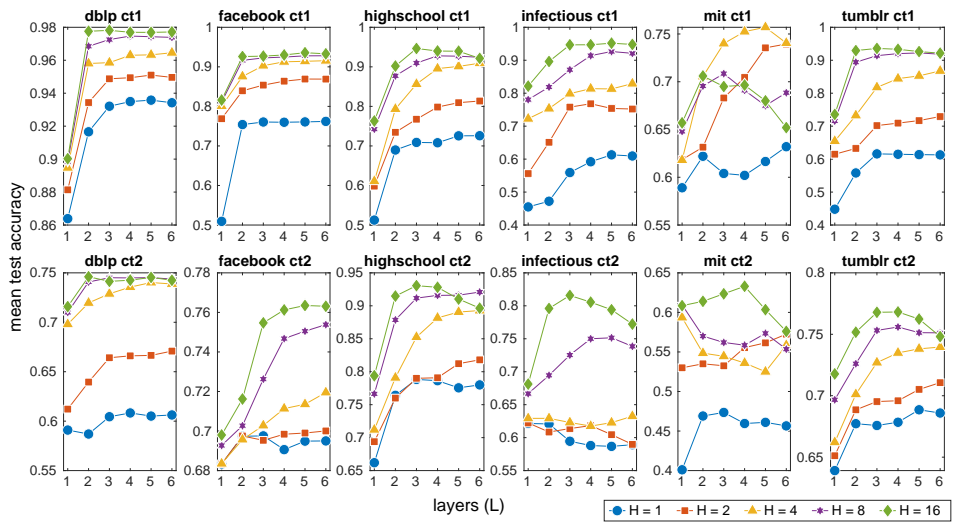}
	\caption{Impact of reservoir layout on classification accuracy.}
	\label{fig:experiments}
\end{figure}

Figure \ref{fig:experiments} shows how both hidden state dimension (i.e. reservoir size per layer) $H \in \{1, 2, 4, 8, 16\}$ and number of layers $L \in 1..6$ affect classification accuracy.
We notice two trends:

(i) increasing the number of hidden units $H$ per layer with fixed depth $L$ increases accuracy, which can be explained by the ability of larger reservoirs to offer richer dynamics;

(ii) increasing the number of layers with fixed $H$ has overall a significant impact on accuracy, with a steep increase up to $L = 3$; the beneficial effect of depth has already been reported for GraphESN \cite{Gallicchio2020}, and in our case could be explained similarly to deep ESN for sequences \cite{Pedrelli2017}, i.e. that deeper layers represent different time-scales of a graph sequence.

We now compare DynGESN against other models.
Temporal graph kernels compute the Gram matrix for a SVM by first transforming the dynamic temporal graphs into static, and then by applying `classic' graph kernels such as $k$-step random walk (RW-$k$) or Weisfeiler--Lehman sub-tree kernel of depth $h$ (WL-$h$).
In \cite{Oettershagen2020} three static transformations are proposed: \emph{reduced graph representation} (RD), which loses temporal information, but does not increase graph dimension; \emph{direct line expansion} (DL) and \emph{static expansion} (SE), which preserve temporal information by providing a much larger static graph representation (Table \ref{tab:comparison}).
Temporal graph kernels APPR-$S$ are also proposed to approximate the random-walk kernel on direct line expansion (i.e. DL-RW) by sampling $k$-step random walks starting on $S$ vertices of the expanded graph.
Since these offer the best trade-off between speed and accuracy on temporal graph kernels, we will consider them in our comparison.

Tables \ref{tab:task1} and \ref{tab:task2} compare the classification accuracy (with standard deviation, computed on bootstraps) of our model for $H = 16$ and $L = 4$ against the results reported in \cite{Oettershagen2020} for temporal graph kernels.
Best accuracies between DynGESN and APPR-$S$ are highlighted in bold; results for exact DL temporal graph kernels are also reported for reference, with SE having similar performances  to DL, and RD significantly poorer with respect to both.

DynGESN performs consistently better than APPR-100, and on par or above APPR-250 on all datasets except from \textsf{mit-ct1} and \textsf{mit-ct2}, which are problematic also for some exact temporal kernel due to their small number of samples.
Notice also that $S$ must scale with dataset size in order to satisfy the theoretical approximation guarantees for APPR-$S$ kernels \cite{Oettershagen2020}.

\begin{table}[h!]\small
	\centering
	\begin{tabular}{l|cccccc}
		\toprule
		\textbf{Model} & \textsf{dblp} & \textsf{facebook} & \textsf{highschool} & \textsf{infectious} & \textsf{mit} & \textsf{tumblr} \\
		\midrule
		DL-RW & $98.7_{\pm 0.1}$ & $96.5_{\pm 0.1}$ & $97.4_{\pm 0.7}$ & $98.0_{\pm 0.4}$ & $92.9_{\pm 0.9}$ & $95.2_{\pm 0.6}$ \\
		DL-WL & $98.5_{\pm 0.2}$ & $96.6_{\pm 0.2}$ & $99.2_{\pm 0.6}$ & $98.1_{\pm 0.4}$ & $91.7_{\pm 1.6}$ & $94.2_{\pm 0.4}$ \\
		\midrule
		APPR-50  & $93.1_{\pm 0.5}$ & $89.4_{\pm 0.5}$ & $87.6_{\pm 1.7}$ & $83.4_{\pm 1.6}$ & $82.8_{\pm 2.0}$ & $89.3_{\pm 0.7}$ \\
		APPR-100 & $95.7_{\pm 0.5}$ & $92.4_{\pm 0.4}$ & $90.2_{\pm 1.9}$ & $91.1_{\pm 1.0}$ & $85.1_{\pm 2.6}$ & $90.4_{\pm 1.0}$ \\
		APPR-250 & $97.2_{\pm 0.2}$ & $\mathbf{94.6}_{\pm 0.3}$ & $94.0_{\pm 1.3}$ & $\mathbf{95.1}_{\pm 0.9}$ & $\mathbf{89.3}_{\pm 2.8}$ & $92.7_{\pm 0.3}$ \\ 
		\midrule
		DynGESN & $\mathbf{97.7}_{\pm 1.8}$ & $93.0_{\pm 2.5}$ & $\mathbf{94.4}_{\pm 5.3}$ & $94.7_{\pm 5.3}$ & $69.7_{\pm 9.2}$ & $\mathbf{93.3}_{\pm 3.9}$ \\
		\bottomrule
	\end{tabular}
	\caption{Classification accuracy on \textsf{ct1} tasks.}
	\label{tab:task1}
\end{table}

\begin{table}[h!]\small
	\vspace*{-1.7em}
	\centering
	\begin{tabular}{l|cccccc}
		\toprule
		\textbf{Model} & \textsf{dblp} & \textsf{facebook} & \textsf{highschool} & \textsf{infectious} & \textsf{mit} & \textsf{tumblr} \\
		\midrule
		DL-RW & $81.8_{\pm 0.9}$ & $80.0_{\pm 0.5}$ & $93.4_{\pm 1.0}$ & $88.7_{\pm 1.2}$ & $82.6_{\pm 2.1}$ & $77.2_{\pm 1.0}$ \\
		DL-WL & $76.5_{\pm 1.0}$ & $80.0_{\pm 0.5}$ & $89.3_{\pm 0.7}$ & $78.7_{\pm 1.5}$ & $36.4_{\pm 4.0}$ & $78.2_{\pm 1.3}$ \\
		\midrule
		APPR-50  & $69.7_{\pm 0.8}$ & $72.0_{\pm 0.7}$ & $76.4_{\pm 3.0}$ & $74.6_{\pm 1.8}$ & $60.4_{\pm 4.0}$ & $74.7_{\pm 1.3}$ \\
		APPR-100 & $74.5_{\pm 0.7}$ & $73.1_{\pm 0.6}$ & $83.8_{\pm 1.8}$ & $75.6_{\pm 1.9}$ & $65.4_{\pm 4.2}$ & $76.5_{\pm 1.5}$ \\
		APPR-250 & $\mathbf{76.4}_{\pm 0.9}$ & $\mathbf{77.9}_{\pm 0.6}$ & $90.4_{\pm 1.8}$ & $78.6_{\pm 2.1}$ & $\mathbf{66.9}_{\pm 2.5}$ & $\mathbf{78.4}_{\pm 1.3}$ \\ 
		\midrule
		DynGESN & $74.3_{\pm 4.7}$ & $76.1_{\pm 3.9}$ & $\mathbf{92.8}_{\pm 5.2}$ & $\mathbf{80.6}_{\pm 9.1}$ & $63.3_{\pm 11.0}$ & $76.8_{\pm 6.2}$ \\
		\bottomrule
	\end{tabular}
	\caption{Classification accuracy on \textsf{ct2} tasks.}
	\label{tab:task2}
\end{table}

DynGESN is able to obtain an embedding for each time-step needing only previous vertex features and current labels and edges, thus requiring $O(V)$ space and $O(V + E_t)$ matrix operation at each time-step ($O(T V + E)$ overall; $\sim 10^{-4}s$ per time-step on a laptop for our implementation).
Thus our model can be applied in an `on-line' setting, updating vertex and graph embedding directly as the dynamic graph evolves, without needing anything more then $\vect{x}_{t-1}$, $\vect{u}_t$ and $\mathcal{E}_t$ at each time-step.
Furthermore, DynGESN can be easily adopted for vertex regression/classification tasks by simply skipping graph pooling \eqref{eq:pooling}.

\begin{table}\small
	\centering
	\begin{tabular}{l@{}cc}
		\multicolumn{3}{c}{\textsc{Processing graph size}} \\
		\toprule
		 & \textbf{vertices} & \textbf{edges} \\
		\midrule
		RD & $O(V)$ & $O(V^2)$ \\
		DL & $O(E)$ & $O(E^2)$ \\
		SE & $O(E)$ & $O(E)$ \\
		\midrule
		DynGESN & $O(V)$ & $O(E_t)$ \\
		\bottomrule
	\end{tabular}
\quad
	\begin{tabular}{lccc}
	\multicolumn{4}{c}{\textsc{Embedding time complexity}} \\
	\toprule
	 & RW-$k$ & APPR-$S$ & WL-$h$ \\
	\midrule
	RD & $O(V^{2 k})$ & --- & $O(N^2 h V^2)$ \\
	DL & $O(E^{2 k})$ & $O(S k)$ & $O(N h E^2 + N^2 h E)$ \\
	SE & $O(E^k)$ & --- & $O(N^2 h E)$ \\
	\midrule
	\multicolumn{4}{l}{DynGESN \quad $O(V + E_t)$ per time-step} \\
	\bottomrule
	\end{tabular}
\vspace*{-0.45em}
	\caption{Comparison of space requirements and time complexity ($V$, $E$ and $E_t$ are the cardinalities of $\mathcal{V}$, $\mathcal{E}$ and $\mathcal{E}_t$, respectively, for an input dynamic graph; $N$ is number of dataset samples).}
	\label{tab:comparison}
\end{table}

\section{Conclusions}
We have presented an extension of static graph echo state networks for the efficient processing of dynamic graphs with time-dependent vertex labels, along with theoretical conditions for the echo state property to hold.
Experiments performed on twelve dissemination process classification tasks show an accuracy comparable to approximate temporal graph kernels paired with SVMs.
These results, along with significantly lower space requirements, the ability to embed graphs \emph{on-line} (thus to amortise the cost on time-steps), and the lack of training for the embedding function, suggest that our method is the most promising for scaling on larger real-world applications.
Finally, our model can also be easily employed for classification/regression tasks on vertices, which will be examined in subsequent works.

\begin{footnotesize}

\bibliographystyle{unsrt}
\bibliography{bibliography}

\end{footnotesize}

\end{document}